\documentclass{article}
\usepackage{spconf,amsmath,graphicx,hyperref,marvosym,multirow,xcolor,colortbl,booktabs,amsfonts,cite}

\usepackage[normalem]{ulem}
\useunder{\uline}{\ul}{}

\definecolor{mygreen}{HTML}{00AA00}
\title{Federated Joint Learning for Domain and Class Generalization}
\twoauthors
 {Haoran Xu\textsuperscript{\ddag}}
 {Zhejiang University\\
  Hangzhou, China}
  {Jiaze Li\textsuperscript{\ddag}\textsuperscript{\dag}, Jianzhong Ju, Zhenbo Luo\textsuperscript{\Letter}}
	{MiLM Plus\\
	Xiaomi Inc.\\
	Beijing, China}

\begin{document}

\ninept
\maketitle
\begin{abstract}
Efficient fine-tuning of visual-language models like CLIP has become crucial due to their large-scale parameter size and extensive pretraining requirements. Existing methods typically address either the issue of unseen classes or unseen domains in isolation, without considering a joint framework for both. In this paper, we propose \textbf{Fed}erated Joint Learning for \textbf{D}omain and \textbf{C}lass \textbf{G}eneralization, termed \textbf{FedDCG}, a novel approach that addresses both class and domain generalization in federated learning settings. Our method introduces a domain grouping strategy where class-generalized networks are trained within each group to prevent decision boundary confusion. During inference, we aggregate class-generalized results based on domain similarity, effectively integrating knowledge from both class and domain generalization. Specifically, a learnable network is employed to enhance class generalization capabilities, and a decoupling mechanism separates general and domain-specific knowledge, improving generalization to unseen domains. Extensive experiments across various datasets show that \textbf{FedDCG} outperforms state-of-the-art baselines in terms of accuracy and robustness.

\end{abstract}
\begin{keywords}
Federated Learning, Domain Generalization, Class Generalization
\end{keywords}

\renewcommand{\thefootnote}{} 
\footnotetext{\ddag~Equal contribution}
\footnotetext{\dag~Project leader}
\footnotetext{\Letter~Corresponding authors}

\section{Introduction}\label{sec:intro}

Recent advancements in visual-language models~\cite{jia2021scalingvisualvisionlanguagerepresentation,li2025imove,li2025multilevelsemanticawaremodelaigenerated,10678147}, epitomized by CLIP~\cite{radford2021learningtransferablevisualmodels}, have demonstrated robust feature extraction through extensive pretraining on large-scale image-text pairs, achieving competitive performance across a multitude of downstream tasks~\cite{Zhou_2022,zhu2024promptalignedgradientprompttuning,xin2023mmapmultimodalalignment,zeng2024focus} such as image classification and text-to-image retrieval. Despite the impressive capabilities of CLIP, fine-tuning this model to improve performance on downstream tasks is frequently constrained by the substantial computational demands due to its large number of parameters. As a result, recent research has increasingly focused on developing efficient fine-tuning methodologies~\cite{hu2021loralowrankadaptationlarge} to address this challenge.
\par
As one of the mainstream effient fine-tuning methods, CoOp~\cite{zhou2022coop} shows great potential. It introduces learnable soft prompts to replace manually crafted ones, enabling efficient fine-tuning of CLIP with limited data, thereby achieving superior performance in downstream tasks. Building on this, FedCoOp~\cite{guo2022promptflletfederatedparticipants} integrates CoOp into federated learning (FL) to facilitate the learning of a prompt set across diverse client datasets while preserving data privacy. However, there are two main types of challenges for existing prompt tuning methods when migrating to FL for downstream applications, i.e., \textbf{\textit{class generalization for unseen class and domain generalization for unseen domain}}.
\par
Previous work on class generation ideas is based on prompt generators which are usually implicit neural networks for class context to enhance generalization of unseen classes, such as FedTPG~\cite{qiu2024federated}. FedTPG implicitly captures class context via a learnable embedding and extends prompt tuning to unseen classes through its class context learning network. Meanwhile, for domain generalization, DiPrompT~\cite{bai2024dipromptdisentangledprompttuning}, is to explicitly decouple knowledge into general and domain-specific segments, modeling these through separate learnable prompts. By explicitly modeling and integrating seen domains, the method is expected to have the ability to understand and model unseen domains.

\par
This is the fact that existing methods have achieved great success in the field of class generalization and domain generalization respectively. However, in real and complex scenarios, there is both unseen class and unseen domain in the test dataset. Current methods typically address only one of these challenges, resulting in their poor performance while the scenario in which both unseen classes and unseen domains coexist has not been thoroughly investigated in prior research.


Built upon the above, we ask one question:
\textbf{\textit{Is there a method that can address the combined challenge of unseen classes and unseen domains in federated learning setting?}}

Inspired by previous work~\cite{10377975,10204197,zhai2023multipromptslearningcrossmodalalignment, xu2025federated,wang2024fedcda,xu2025federatedlearningsamplelevelclient,li2026federatedbalancedlearning,xu2026visionneednavigatingoutofdistribution},we propose \textbf{FedDCG}, \textbf{\textit{a novel approach that addresses both class and domain generalization in federated learning setting based on decoupling concept}}. Specifically, during the training period, we 
adopt Class-Specific Domain-Grouping Collaborative Training Strategy to train the model to solve class generalization and domain generalization issues. In simple terms, we group domains and train a class-generalized network within each group to avoid decision boundary confusion in classes and domains. During inference, we aggregate the class-generalized results based on domain similarity, thereby integrating knowledge from both class and domain generalization.
Our contributions can be summarized as follows:
\begin{itemize}
\item We are the first to propose a novel task setting combined challenge of unseen classes and unseen domains.

\item We introduce \textbf{FedDCG}, which groups domains and trains class-generalized networks within each group to prevent decision boundary confusion, integrating class and domain generalization knowledge through domain similarity during inference.

\item Our method consistently outperforms current state-of-the-art baselines across extensive experiments, demonstrating exceptional robustness.
\end{itemize}

\begin{figure*}[t]
\centering
\includegraphics[width=0.62\linewidth]{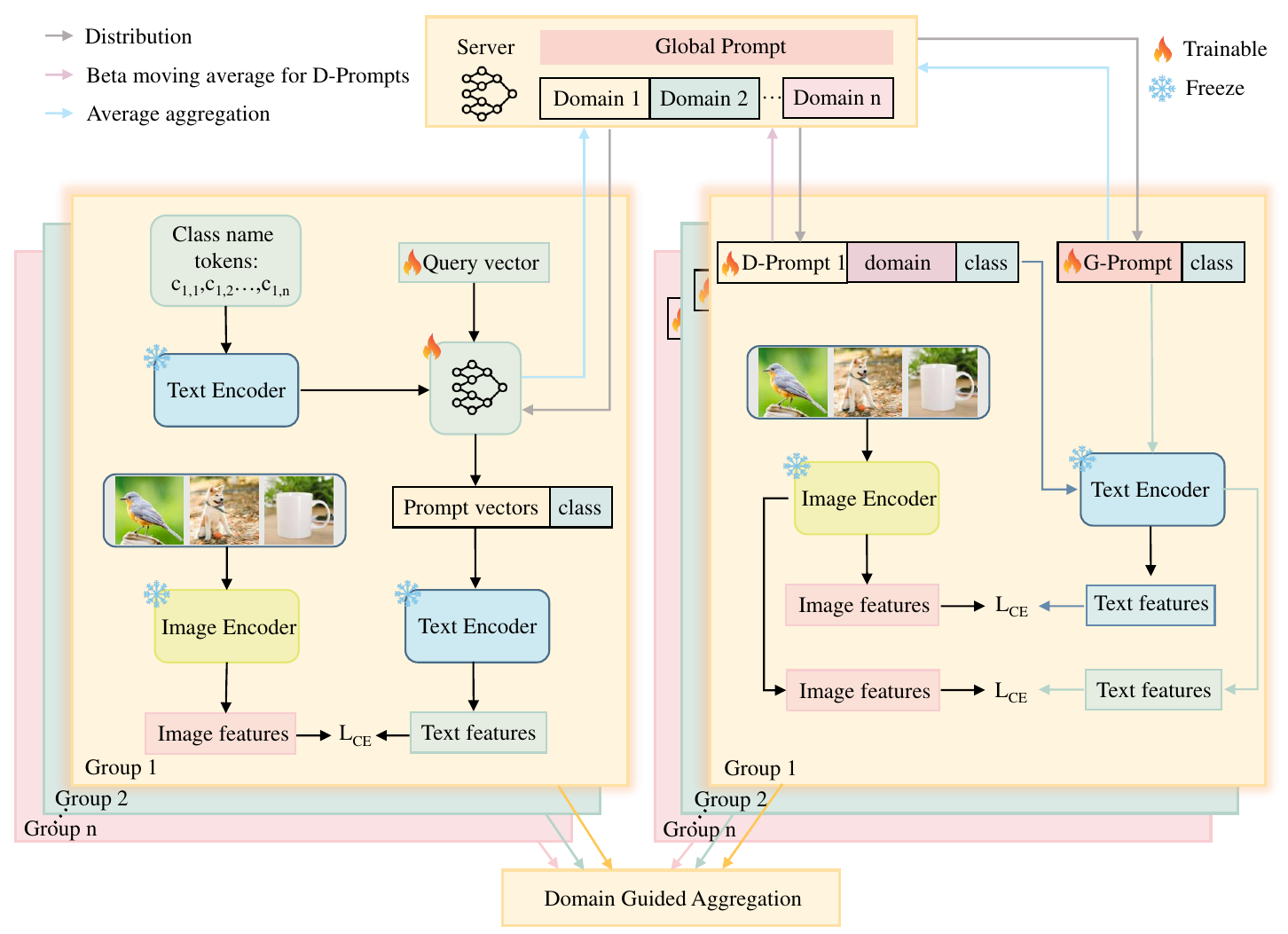}
\caption{Illustration of our method. Our approach is a dual strategy: on the left, we group domains to learn class generalization networks, while on the right, we perform domain disentanglement pretraining.}
\label{fig:Pipeline} 
\end{figure*}

\section{Method}\label{sec:method}

In this section, we introduce \textbf{FedDCG} in detail which is proposed to address the combined challenge of unseen classes and unseen domains in fer. Drawing inspiration from prior research ~\cite{10377975,10204197,zhai2023multipromptslearningcrossmodalalignment}, we adopted a strategy of segregating the dataset into domain-specific groups. Within each of these groups, a class generalization network is learned independently, thereby avoiding the entanglement of decision boundaries in classes and domains and achieving collaborative optimization. \textbf{\textit{Our approach is divided into three parts: Domain-based Grouping, Class-Specific Domain-Grouping Collaborative Training and Domain-Guided Aggregation Inference}}. Specifically, our method firstly use Domain-based Grouping to group networks by domains. Then \textbf{FedDCG} uses Class-Specific Domain-Grouping Collaborative Training strategy to train model. Finally the results are obtained by Domain Guided Aggregation Inference method. It is formulated as figure \ref{fig:Pipeline}.

\subsection{Domain-based Grouping}
In this section, we mathematically formulate the grouping method as follows:
\begin{equation}
f_{\theta_i}(T) = h_{\phi}\!\left(\text{CrossAttention}\!\left(Q, K_{T_i}, V_{T_i}\right)\right), \quad i = 1, \dots, m
\end{equation}
Here, $m$ denotes the total number of distinct domains. For each domain $D_i$, the function $f_{\theta_i}(T)$ represents a class generalization model that leverages cross-attention over the task-specific keys $K_{T_i}$ and values $V_{T_i}$, conditioned on the query $Q$. The transformation $h_{\phi}(\cdot)$ further refines the attended representation to produce a domain-adapted output. In this stage, we initialize the learnable embedding, which will be used for the subsequent collaborative alternating training.

\subsection{Class-Specific Domain-Grouping Training}
In this part, we divide the following training stage into two sub-stages: Class-specific Grouping Training and Domain-specific Decoupling Training. These two sub-stages will proceed alternately in coordination.

\subsubsection{Class-Specific Domain-Grouping Training}
We firstly introduce the Class-specific Grouping Training. This sub-stage will fix the domain embedding and focus on training the model to solve the class generalization problem. Its objective is to optimize the parameters of the network \( f_{\theta_i}(T) = h_{\phi}(\text{CrossAttention}(Q, K_{T}, V_{T})) \), where \(T\) represents the task-specific text embeddings, and \( h_{\phi}\) is the final output layer. Initially, class names are embedded into text embeddings using a pretrained text encoder, represented as \( T = \{E_{text}(c_j)\}_{j=1}^n \), where \( c_j \) corresponds to each class in Domain \(i\).

To achieve effective generalization, the cross-attention mechanism is employed within the network to derive context-aware prompt vectors. Specifically, the cross-attention module computes key and value pairs from the text embeddings \( K_T = T \times W_K \) and \( V_T = T \times W_V \), where \( W_K \) and \( W_V \) are learnable weight matrices. The query vector \( Q \) interacts with these pairs to generate the attention output, which is further processed by the network \( g_{\psi} \) to produce the final prompt vectors.

The network is trained to minimize the prediction loss on the class labels of Domain \(i\). The prediction probability for an input image \( x \) is defined as:

\begin{equation}
p_{\theta}(y = j \mid x, T) = \frac{\exp\left(\frac{1}{\tau} \cdot s_j \right)}{\sum_{k=1}^n \exp\left(\frac{1}{\tau} \cdot s_k \right)}
\end{equation}

\begin{equation}
s_j = \text{cos}(E_{\text{img}}(x), E_{\text{text}}(t_j))
\end{equation}
Here, \( \tau \) is a temperature scaling parameter, and \( t_j \) is the concatenation of the prompt vector and class name embedding for class \( j \). The loss function is defined as:

\begin{equation}
\mathcal{L}(\theta; T) = - \mathbb{E}_{(x, y) \sim D_i} \left[ y \cdot \log p_{\theta}(y | x, T) \right]
\end{equation}

where \( D_i \) represents the dataset corresponding to Domain \(i\). After training the class generalization network on the client, the model is uploaded to the server. Networks trained on the same domain are aggregated through averaging, enabling cross-client collaborative training.

\subsection{Domain-specific Decoupling Training}
In the second sub-stage, we will fix the class generalization network and focus on the optimization of domain embedding, in order to solve the domain generalization problem. Our approach employs a domain-specific decoupling strategy, inspired by DiPrompT, to separate general and domain-specific knowledge across federated clients. This method involves Global Prompt (\( P_G \)) and Domain Prompts (\( P_D \)). The global prompt captures domain-invariant features, while domain prompts extract domain-specific knowledge. Both prompts are trained concurrently, with the global prompt optimized for cross-client generalization, and domain prompts tailored to retain distinct domain features.
\par
We apply two key aggregation strategies from DiPrompT to ensure effective knowledge transfer and stability:

\textbf{Domain-Wise Aggregation Strategy.} The domain prompts from clients sharing the same domain are aggregated using a weighted average:
\begin{equation}
V_D^{s+1} = V_D^s + \frac{\sum_{i=1}^{M} |D_i| \cdot \Delta V_D^{s+1}}{\sum_{i=1}^{M} |D_i|},
\end{equation}
where \( \Delta V_D^{s+1} \) represents the client updates, and \( |D_i| \) is the dataset size on the \( i \)-th client. This ensures that the aggregated prompt effectively consolidates domain-specific features.

\textbf{Beta Momentum Averaging Mechanism.} To stabilize learning and reduce client drift, we employ beta momentum averaging, which updates the domain prompt \( V_D \) as:
\begin{equation}
\tilde{V}_D^{s+1} = \frac{\sum_{j=0}^{s} \alpha_j V_D^j}{\sum_{j=0}^{s} \alpha_j} + \alpha_s V_D^{s+1},
\end{equation}
where \( \alpha_j \) is derived from a beta distribution, ensuring balanced contributions from past and current updates.

The above two sub-stages will proceed concurrently and alternately, jointly achieving Class-Specific Domain-Grouping Training Strategy.

\begin{table*}[ht]
	\centering
    \resizebox{\linewidth}{!}{
	\begin{tabular}{lccccccccccccccc}
		\toprule
        & & \multicolumn{5}{c}{ImageNet-R} & \multicolumn{5}{c}{ImageNet-A} \\
        \cmidrule(r){1-1} \cmidrule(r){2-2} \cmidrule(r){3-7} \cmidrule(r){8-12}
		Datasets & Method &\multicolumn{1}{c}{FedCoOp} & \multicolumn{1}{c}{FedCoCoOp} &
  \multicolumn{1}{c}{FedTPG} & 
  \multicolumn{1}{c}{DiPrompT} & \multicolumn{1}{c}{Ours} &\multicolumn{1}{c}{FedCoOp} & \multicolumn{1}{c}{FedCoCoOp} & \multicolumn{1}{c}{FedTPG} & 
  \multicolumn{1}{c}{DiPrompT} & \multicolumn{1}{c}{Ours}\\ 
		\cmidrule(r){1-1} \cmidrule(r){2-2}
  \cmidrule(r){3-7}\cmidrule(r){8-12}  
		\multirow{5}{*}{Minidomainnet} & 
Painting & 64.15 & 65.48  & 65.17 & 66.38 & \textbf{67.13}  & 52.29 & 53.67 & 54.16 & 55.28& \textbf{57.15} \\
		& Clipart & 62.14 & 64.26  & 64.89 & 65.17 & \textbf{66.80}  & 54.15 & 55.23 & 56.87 & 56.26& \textbf{57.12} \\ 
       & Sketch & 60.03 & 61.24  & 61.11 & 63.14 & \textbf{64.89}  & 50.11 & 52.19 & 54.17& 54.82& \textbf{55.67} \\ 
       & Real & 67.46 & 67.93  & 68.32 & 69.43 & \textbf{70.18}  & 57.85 &   58.36 & 59.11& 60.35& \textbf{61.48} \\ 
       & Average & 63.44 & 64.72  & 64.87 & 66.03 & \textbf{67.23}  & 53.60 & 54.86 & 56.08& 56.67& \textbf{57.86} \\ 
		\midrule
        \multirow{5}{*}{OfficeHome} & Art & 57.12 & 58.12& 58.39  & 61.58 & \textbf{66.24} & 41.73  & 42.85  & 42.51 & 44.15 & \textbf{45.29} \\
		& Clipart & 57.68  & 57.94 & 58.44  & 59.39 & \textbf{62.59} & 40.71  & 41.51  & 41.79 & 43.18 & \textbf{44.79} \\ 
       & Product & 64.26  & 64.76 & 65.87  & 73.10 & \textbf{74.57} & 36.58  & 38.06  & 38.93 & 42.65 & \textbf{44.24} \\ 
       & Real & 66.48  & 67.85 & 68.89  & 75.74 & \textbf{77.18} & 38.44  & 39.67 & 39.76 & 44.68 &\textbf{46.10} \\ 
       & Average  & 61.38 & 62.17 & 62.90  & 67.45 & \textbf{70.30} & 39.36  & 40.52  & 40.75 & 43.67 & \textbf{45.11} \\ 
		\bottomrule
	\end{tabular}}
    \caption{Comparison between our method and the baselines under different datasets on the combined unseen classes and unseen domains setting. \textbf{Bold} fonts highlight the best accuracy.}
     \label{result1}
\end{table*}

\subsection{Domain Guided Aggregation Inference}
In this stage, we implement a domain-decoupling strategy that simultaneously achieves class and domain generalization, even for unseen domains. Specifically, the model utilizes the overall similarity score \( w_m \) of a sample to each domain to weight the inference results of the corresponding class generalization network. The output of the class generalization network for the \( m \)-th domain is represented by a vector \( Z_{mj} \), which is a concatenation of the output features from the domain-specific class generalization network and the class-specific features. Mathematically, this can be expressed as:

\begin{align}
P(y = j | x) &= \sum_{m=1}^M w_m \cdot \text{sim}(I(x), Z_{mj}) + \\
&\quad w_g \cdot \text{sim}(I(x), Z_j) \nonumber
\end{align}
where \( Z_{mj} = \text{concat}(f_{\theta_m}(x), t_j) \) represents the concatenated vector of the \( m \)-th domain's class generalization network output \( f_{\theta_m}(x) \) and the class feature \( t_j \). Here, \( Z_j \) denotes the global prompt vector concatenated with the class feature vector, representing the textual features used globally across domains.

\section{Experiments}\label{sec:exp}

\subsection{Experiment Setting} \label{experiment_settings}
\subsubsection{\textbf{Dataset.}} 
Following prior works \cite{xin2023mmapmultimodalalignment,bai2024dipromptdisentangledprompttuning},we design and utilize the Office-Home \cite{venkateswara2017Deep} and MiniDomainNet \cite{zhang2023domainadaptor} datasets for training, as these datasets are particularly suitable for evaluating both class generalization and domain generalization in a federated learning setting. To ensure that our model is exposed to a diverse range of classes and domains during training, we incorporate ImageNet-Rendition \cite{5206848} and ImageNet-A \cite{5206848} datasets for zero-shot evaluation. This approach allows us to rigorously assess the model's ability to generalize across unseen classes and domains.
\begin{table}[ht]
	\centering
    \resizebox{\linewidth}{!}{
	\begin{tabular}{lcccccccccc}
		\toprule
		\multicolumn{1}{c}{Aggregation Method} & Average& Uncertainty&Domain Guided & Result \\
		\midrule
		M1 &\checkmark &  & &  73.21 \\
        M2 & & \checkmark & &  73.14 \\
        M3(Ours) & &  &\checkmark & \textbf{74.57}  \\
		\bottomrule
	\end{tabular}}
    \caption{Ablation Study. The setting is training on OfficeHome and test on ImageNet-R.}
    \label{result4}
    \vspace{-0.5cm}
\end{table}
\subsubsection{\textbf{Model Architecture.}} 
All experiments in this study are conducted using the ViT-B/16 \cite{dosovitskiy2021imageworth16x16words} variant of the CLIP \cite{radford2021learningtransferablevisualmodels} model as the backbone. 
In line with the FedTPG configuration, our class generalization network incorporates a 4-head cross-attention layer, with each layer having a hidden dimension of 512. This design allows for effective communication between different modalities and captures nuanced relationships between global and domain-specific information. 



\subsubsection{\textbf{Implementation Details.}}
In our experiments, each model was trained for 250 rounds, with training domains selected from three distinct domains in both datasets. We allocated 18 clients for training in each case. Given the variation in the number of categories between MiniDomainNet and Office-Home, we selected 20 classes per client for MiniDomainNet and 10 classes per client for Office-Home. Our approach involves initially training a domain-specific class generalization network, followed by domain-decoupled training. We employed an SGD optimizer with a learning rate of 1e-3, utilizing cosine annealing for learning rate decay. The batch size was set to 128. 


\textbf{Baselines.}
In order to conduct a comprehensive evaluation, we compare our proposed method with with the following state-of-the-art methods, es, including: FedCoOp\cite{guo2022promptflletfederatedparticipants}, FedCoCoOp\cite{zhou2022conditionalpromptlearningvisionlanguage}, FedKgCoOp\cite{kgcoop23}, FedCoPL\cite{goswami2023coplcontextualpromptlearning},FedMaple\cite{khattak2023maplemultimodalpromptlearning}, FedTPG\cite{qiu2024federated}, DiPrompT\cite{bai2024dipromptdisentangledprompttuning}.

\subsection{Performance Evaluation}
\par
We demonstrate the best mean accuracy of each baseline and the proposed method in Table~\ref{result1}.


\par
We can find that our method outperforms all the baseline on test dataset ImageNet-R and ImageNet-A.
Meanwhile, we find that the performance of DiPrompT is better than that of FedTPG, which indicates that the decision boundary of purely implicit methods will be confused under the cooperative setting of class generalization and domain generalization. Meanwhile, the table also shows the results of training and testing in different domains. We can find that under the two datasets, when the real domain is the test domain, i.e., the other three domains are the training domain, the performance is the highest, which indicates that the real domain is easier to generalize and learn than other domains. In contrast, clipart and art domain are more difficult to learn and more difficult to generalize. When we compare the results of different test sets, we can find that the test results under ImageNet-R tend to be higher and the test results under ImageNet-A tend to be lower, which means that generalization of Image-A is more difficult than generalization of ImageNet-R. In the case of training on the Office-Home dataset and testing on the ImageNet-R test set, our method on average synthesized nearly 3\% higher than the second-place method, DiPrompT.


\subsection{Low sampling}
In this section, we test our approach against different baselines at a lower sampling rate and pick Office-Home dataset as the training dataset. Meanwhile, when we set the sampling rate to 50\%, it means that we use less training data of different domains to test under the same data. We test the effectiveness of our method under this more difficult domain generalization and class generalization setting. As shown in Table~\ref{result2} and Table~\ref{sub:result3}, our approach still maintains best results in most cases. Specifically, our average performance across two datasets and four domains exceeded other baseline approaches. In fact, the effect of our method is only lower than that of DiPrompT. In addition, FedTPG's method shows more degradation, indicating that the explicit method DiPrompT is significantly more effective, convergent and generalizing than the implicit method with less data sampling rate and less training data. Furthermore, at a lower sampling rate, the performance of real domain as the test domain is more different from that of other domian domains as the test domain, which indicates that it is more difficult to generalize Art,Clipart and Product domain under the condition of lower sampling rate.

\subsection{More Analysis}
\begin{table}[ht]
	\centering
    \resizebox{\linewidth}{!}{
	\begin{tabular}{lccccccc}
		\toprule
  & & \multicolumn{5}{c}{Office-Home(\%)} \\ 
		\cmidrule(r){1-1} \cmidrule(r){2-2}
        \cmidrule(r){3-7}
		\multicolumn{1}{c}{Datasets}&\multicolumn{1}{c}{Method} & Art & Clipart  & Product & Real & Average \\
		\midrule
		\multirow{7}{*}{ImageNet-R}&FedCoOp & 45.12 & 45.48 & 42.09 & 49.84 &45.63  \\
&FedCoCoOp & 45.67 & 46.17 & 42.35 & 49.31 &45.87  \\
&FedKgCoOp & 46.12 & 45.98 & 42.79 & 50.55 &46.36\\
&FedCoPL & 43.29 & 44.02 & 41.06 & 47.66  &44.00\\
&FedMaple & 44.29 & 45.97 & 42.78 & 49.34 & 45.17 \\
&FedTPG & 46.68 & 45.68 & 43.82 & 51.69  &46.97\\
&DiPrompT & 47.11 & \textbf{46.78} & 45.16 & 52.31&47.84  \\
&Ours & \textbf{47.48} & 46.44 & \textbf{45.78} &  \textbf{52.97} &\textbf{48.17} \\
\midrule
\multirow{7}{*}{ImageNet-A}&FedCoOp & 43.26 & 40.39 & 39.44 & 46.55 &42.41 \\
&FedCoCoOp & 43.19 & 41.15 & 40.87 & 46.85 & 43.01 \\
&FedKgCoOp & 44.39 & 41.87 & 40.56 & 46.79 & 43.40 \\
&FedCoPL & 42.53 & 41.38 & 40.76 & 45.21 & 42.47 \\
&FedMaple & 43.16 & 42.37 & 40.58 & 47.11 & 43.30 \\
&FedTPG & 44.88 & 42.97 & 41.19 & 48.94 & 44.49 \\
&DiPrompT & 45.73 & 43.86 & 42.33 & \textbf{49.32} &  45.31 \\
&Ours & \textbf{46.41} & \textbf{45.22} & \textbf{43.58} & 48.98 &\textbf{46.05} \\
		\bottomrule
	\end{tabular}}
    \caption{Comparison with  baselines under low sampling rate on Office-Home dataset. \textbf{Bold} fonts highlight the best accuracy.}
    \label{result2}
    \vspace{-0.2cm}
\end{table}

\begin{table}[ht]
	\centering
    \resizebox{\linewidth}{!}{
	\begin{tabular}{lccccccc}
		\toprule
  & & \multicolumn{5}{c}{MiniDomainnet(\%)} \\ 
		\cmidrule(r){1-1} \cmidrule(r){2-2}
        \cmidrule(r){3-7}
		\multicolumn{1}{c}{Datasets}&\multicolumn{1}{c}{Method} & Painting & Clipart  & Sketch & Real & Average \\
		\midrule
		\multirow{7}{*}{ImageNet-R}&FedCoOp & 48.43 & 48.95 & 41.35 & 49.12 &46.96  \\
&FedCoCoOp & 49.35 & 48.59 & 42.33 & 49.87 &47.53  \\
&FedKgCoOp & 48.64 & 49.37 & 43.95 & 50.34 &48.07\\
&FedCoPL & 48.17 & 50.04 & 42.58 & 50.38 &47.79\\
&FedMaple & 46.27 & 47.31  & 41.79 & 48.34 &45.92 \\
&FedTPG & 51.66 & 51.15 & 45.34 & 51.64 &49.94\\
&DiPrompT & 52.13 & \textbf{51.77} & 52.08 & 52.31&52.07  \\
&Ours & \textbf{53.04} & \textbf{52.39} & \textbf{53.58} &  \textbf{53.11} &\textbf{53.03} \\
		\bottomrule
	\end{tabular}}
        \caption{Comparison with baselines under low sampling rate on MiniDomainnet dataset. \textbf{Bold} fonts highlight the best accuracy.}
	\label{sub:result3}
    \vspace{-0.4cm}
\end{table}

  

\textbf{Analysis of Domain Guided Aggregation.}

In this section, We explored the meaning of Domain Guided Aggregation. We compare it with average aggregation and uncertainty aggregation. The result is shown in table \ref{result4}. We can find that the method using Domain Guided Aggregation achieves the best result. This proves the validity of ours. More result are provided in the Appendix.

\section{Conclusion}\label{sec:conclusion}
In the paper, we firstly propose \textbf{FedDCG} that groups domains and trains a class-generalized network within each group to avoid decision boundary confusion in the federated learning setting. It has both class generalization and domain generalization capabilities. Finally, we verify the proposed method over various datasets, demonstrating robustness.

\vfill\pagebreak




\bibliographystyle{IEEEbib}
\bibliography{refs.bib}

\end{document}